\documentclass[runningheads]{llncs}
\usepackage[T1]{fontenc}
\usepackage{graphicx}
\newcommand{\msn}{Meta-SN}

\usepackage{marvosym}

\usepackage{pifont}

\usepackage{subfig}
\usepackage{multirow}

\usepackage{amsmath, amsfonts}
\usepackage{graphicx}

\usepackage{algorithmicx}
\usepackage{algorithm}
\usepackage{algpseudocode}

\makeatletter
\newcommand{\removelatexerror}{\let\@latex@error\@gobble}
\makeatother

\newcommand{\tabincell}[2]{
\begin{tabular}{@{}#1@{}}#2\end{tabular}
}

\algnewcommand{\LeftComment}[1]{\State \(\triangleright\) #1}

\usepackage{xcolor}

\begin{document}
\title{Meta-Learning Siamese Network for Few-Shot Text Classification}
%
%

\author{Chengcheng Han\inst{1} 
\and Yuhe Wang\inst{1}
\and Yingnan Fu\inst{1}
\and Xiang Li\inst{1}\textsuperscript{(\Letter)}
\and Minghui Qiu\inst{2}
\and Ming Gao\inst{1,3}
\and Aoying Zhou\inst{1}
}
\authorrunning{C. Han et al.}
%
\institute{School of Data Science and Engineering, East China Normal University,
Shanghai, China \\
\email{\{52215903007, 51205903068, 52175100004\}@stu.ecnu.edu.cn}\\
\email{\{xiangli, mgao, ayzhou\}@dase.ecnu.edu.cn}\\
\and
Alibaba Group, Hangzhou, China \\
\email{minghui.qmh@alibaba-inc.com}
\and
KLATASDS-MOE, School of Statistics, East China Normal University, Shanghai, China}

\maketitle              
%

\begin{abstract}
Few-shot learning has been used
to tackle the problem of label scarcity in text classification,
of which 
meta-learning based methods have shown to be effective,
such as
the prototypical networks~(PROTO).
Despite the success of PROTO, 
there still exist three main problems: 
(1) ignore the randomness of the sampled support sets when computing prototype vectors; 
(2) disregard the importance of labeled samples; 
(3) construct meta-tasks in a purely random manner.
In this paper, 
we propose a Meta-Learning Siamese Network,
namely, \emph{\msn}, to address these issues. 
Specifically,
instead of computing prototype vectors from the sampled support sets,
\msn\ utilizes external knowledge (e.g. class names and descriptive texts) for class labels,
which is encoded as the low-dimensional embeddings of prototype vectors.
In addition,
\msn\ presents a novel sampling strategy for constructing meta-tasks,
which gives higher sampling probabilities to hard-to-classify samples.
Extensive experiments are conducted on six benchmark datasets to show the clear superiority of \msn\ over other state-of-the-art models. 
For reproducibility, 
all the datasets and codes are provided at \url{https://github.com/hccngu/Meta-SN}.

\end{abstract}

\section{Introduction}
\label{Introduction}

Text classification is a pivotal task in natural language processing, 
which aims to predict labels or tags
for 
textual units (e.g., sentences, queries, paragraphs and documents).
It has been 
widely used in various downstream applications,
such as Relation Extraction~\cite{RE_wu2017adversarial} and Information Retrieval~\cite{IR_mitra2017neural}.
With the rapid development of deep learning, 
these approaches generally require massive labeled data
as training set,
which is manually expensive to derive.
\begin{figure}[!t]
\centering
\subfloat[PROTO]{
    \label{fig:proto_bias}
    \includegraphics[scale=0.43]{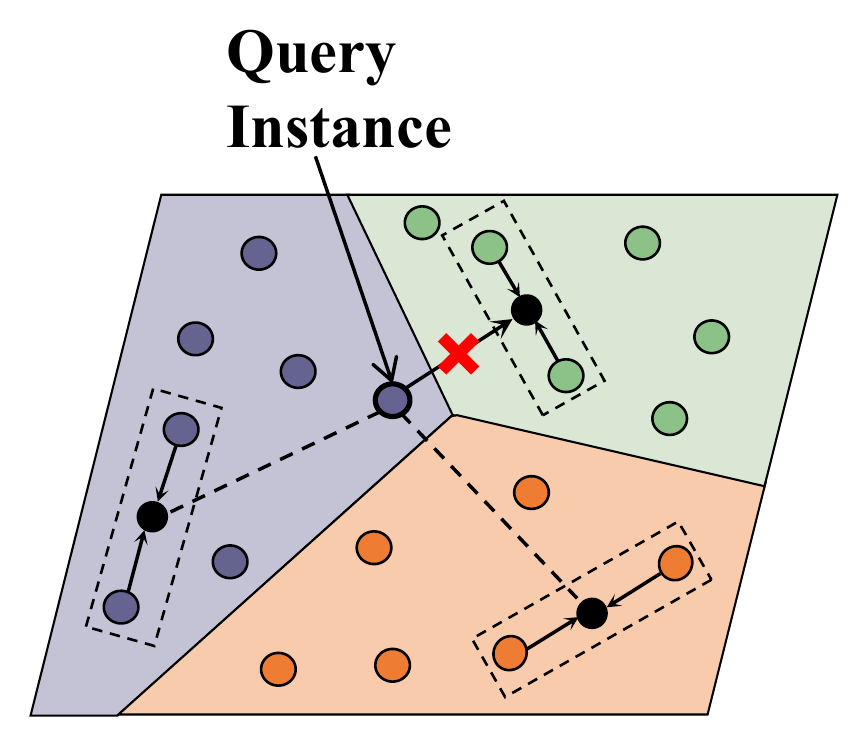}
    }
\subfloat[\!\!Initialization~in~\msn]{
    \label{fig:our_1}
    \includegraphics[scale=0.43]{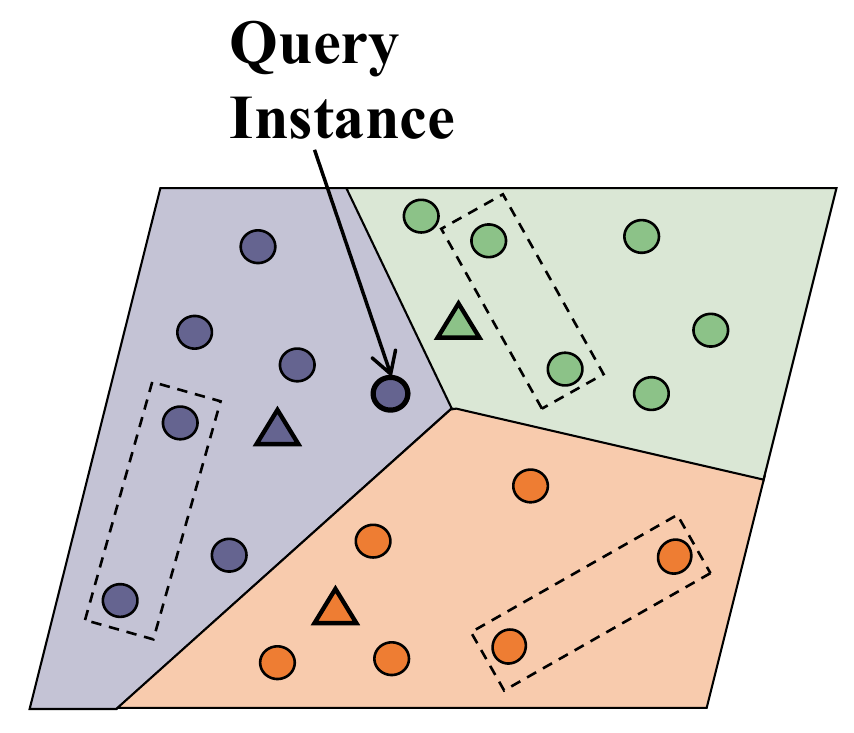}
    }
\subfloat[\small{Refinement in \msn}]{
    \label{fig:our_2}
    \includegraphics[scale=0.43]{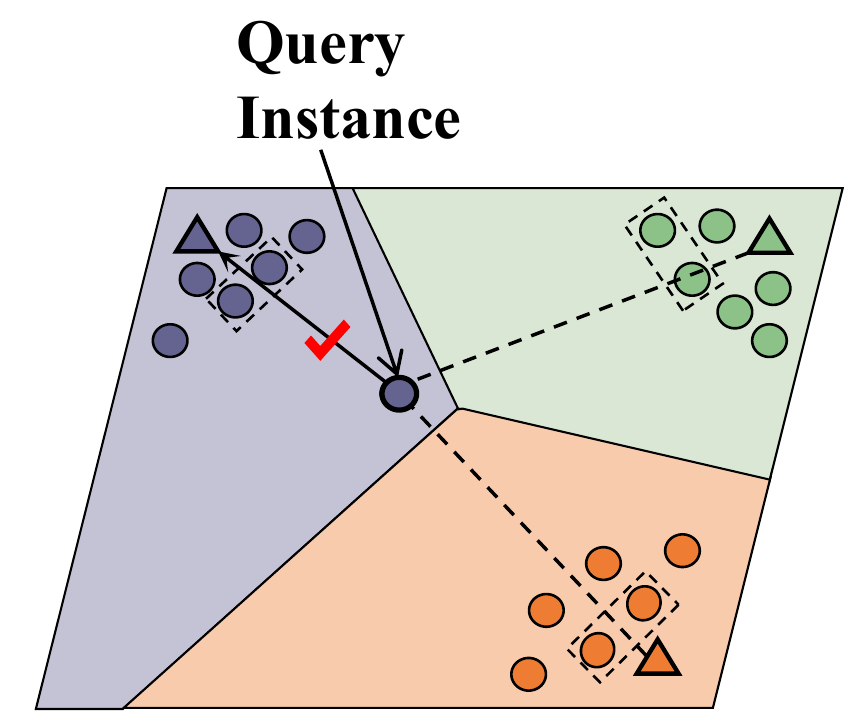}
    }
\caption{A comparison between PROTO and our method~(\msn) in a 3-way 2-shot classification task. 
The dashed box represents the support set randomly sampled from the training data.
\textbf{Left}: 
The black dot represents the prototype vector calculated from the support set. 
Since
the given query instance whose true label is purple is closest to the estimated prototype vector of the green class,
PROTO misclassifies it to green. 
This essentially attributes to the randomness of the sampled support sets.
\textbf{Middle}: 
The triangles represent the initialized prototype vectors computed from the external descriptive texts of classes,
which are independent to the sampled support sets.
\textbf{Right}: 
After refinement with the Siamese network,
prototype vectors and samples are mapped into a low-dimensional space,
where the inter-class distance between different prototype vectors are enlarged and 
the intra-class distance between samples and the corresponding prototype vector is shortened.
}
\end{figure}
To address the problem, 
few-shot learning~\cite{wang2020generalizing} has been proposed,
which aims to train classifiers with scarce labeled data.
Previous studies~\cite{vinyals2016matching_network,Relation_network_sung2018learning,MAML_finn2017model} have shown that meta-learning techniques can be effectively used in
few-shot learning.
In meta-learning, 
the goal is to train a model based on different meta-tasks constructed from the training set
and generalize the model to classify samples in unseen classes from the test set.
Each meta-task contains a \emph{support set} and a \emph{query set}.
Specifically,
the support set is similar to the training set
in traditional supervised learning 
but it only contains a few samples (instances)~\footnote{We interchangeably use sample and instance in this paper.}; 
the query set acts as the test set 
but it can be used to compute gradients for updating model parameters in the training stage. 
As a representative 
meta-learning method, 
the prototypical network (PROTO)~\cite{PROTO_snell2017prototypical} first generates a \emph{prototype vector} for each class by
averaging the embeddings of samples in the support set of the class.
Then it
computes the distance between a query instance in the query set and these prototype vectors.
Finally,
it predicts the query instance to the class with the smallest distance.

Despite the success,
there are three main problems in PROTO.
First,
the true prototype vector of each class should be intuitively fixed.
However, 
the computation of prototype vectors could be adversely affected by the randomness of the sampled support sets,
which could lead to the incorrect prediction of queries' labels (see Figure~\ref{fig:proto_bias}).
Second, 
when calculating prototype vectors,
all the samples in the support set are given the same weight.
This fails to distinguish the 
importance of
samples
when predicting query instances' labels.
Third, 
meta-tasks are randomly constructed from the training data.
This could
lead to simple meta-tasks composed of samples that are easy to be classified,
which are thus insufficient to generalize the model.

In this paper,
to address the issues, 
we propose a \textbf{Meta}-Learning \textbf{S}iamese \textbf{N}etwork, namely, \textbf{\msn}. 
Instead of estimating prototype vectors from the sampled support sets,
we compute these vectors by
utilizing external knowledge on 
the class labels (see Figure~\ref{fig:our_1}),
which
includes class names and 
related 
descriptive texts
(e.g., \emph{Wiki titles} and \emph{Wiki texts}) as shown in Table~\ref{tab example}.
This eliminates the dependence of 
prototype vector estimation on 
the sampled support sets
and also the adverse impact of randomness.
After that,
we further refine these prototype vectors with a Siamese Network.
In particular,
we map both samples and prototype vectors into a low-dimensional space, 
where the inter-class distance between different prototype vectors is enlarged and the intra-class distance between samples and their corresponding prototype vectors is shortened (see Figure~\ref{fig:our_2}).
Further,
we learn the importance of a sample in the support set based on its average distance to the query set.
The closer a sample is to the query set,
the more important the sample is for label prediction,
and
the larger the weight should be assigned.
Finally,
we adopt a novel sampling strategy to construct meta-tasks which assigns higher sampling probability to the hard-to-classify samples.
On the one hand,
the closer the distance between different prototype vectors, 
the more difficult the corresponding classes can be separated.
On the other hand,
the more distant a sample in the support set is to the prototype vector,
the more difficult the classification task will be.
Therefore, 
we give higher sampling probabilities to hard-to-classify tasks to help generalize our model.
The main contributions
of the paper are summarized as follows:
\begin{itemize}
    \item
    We propose a novel \textbf{Meta}-learning \textbf{S}iamese \textbf{N}etwork (\textbf{\msn})\ for few-shot text classification.
    Instead of estimating prototype vectors from
    the sampled support sets,
    \msn\ constructs the prototype vectors with 
    the external descriptive information of class labels
    and further refines these vectors with a Siamese network.
    This alleviates the adverse impact of sampling randomness.

    
    \item We present an effective sampling strategy to construct meta-tasks, which assigns higher sampling probability to the hard-to-classify samples. 
    This boosts the model's generalization ability.
    We further learn the importance of a labeled sample by considering its average distance to the query set.
    
    \item We evaluate the performance of our model on six benchmark datasets, including five text classification datasets and one relation classification dataset. 
    Experimental results demonstrate that \msn\ can achieve significant performance gains over other state-of-the-art methods.
\end{itemize}

\begin{table*}[htbp]
\caption{An example for 3-way 2-shot text classification on the Huffpost dataset, where only two support instances are given in each of the three classes. The ground-truth label of the query instance is Class B. 
External
knowledge
on class labels includes class names and related descriptive texts from Wikipedia, 
which are used to generate prototype vectors in \msn.}
\begin{center}
\resizebox{1\columnwidth}{!}{
\begin{tabular}{|c|l|}
\hline
\multicolumn{2}{|c|}{\textbf{Support set}} \\
\hline
(A) \emph{Politics} & \tabincell{l}{(1) Trump's Crackdown On Immigrant Parents Puts More Kids In An Already Strained System.\\ (2) Ireland Votes To Repeal Abortion Amendment In Landslide Referendum.} \\
\hline
(B) \emph{Entertainment} & \tabincell{l}{(1) Hugh Grant Marries For The First Time At Age 57. \\ (2) Mike Myers Reveals He'd `Like To' Do A Fourth Austin Powers Film.}  \\
\hline
(C) \emph{Sports} & \tabincell{l}{(1) U.S. Olympic Committee Ignored Sexual Abuse Complaints Against Taekwondo Stars: Lawsuit. \\ (2) MLB Pitcher Punches Himself In Face Really Hard After Blowing Game.}  \\
\hline
\multicolumn{2}{|c|}{\textbf{Query instance}}  \\
\hline
Which class? & `Crazy Rich Asians' Trailer Is Already A Magnificent Masterpiece.  \\
\hline
\multicolumn{2}{|c|}{\textbf{External knowledge (class name and related descriptive texts)}}  \\
\hline
(A) & \textbf{Politics} is the set of activities that are associated with making decisions in groups. \\
\hline
(B) & \textbf{Entertainment} is a form of activity that holds the attention and interest of an audience. \\ 
\hline
(C) & \textbf{Sports} pertain to any form of competitive physical activity or game. \\
\hline
\end{tabular}
}
\label{tab example}
\end{center}
\end{table*}

\section{Related Work}
\label{sc:rw}

The mainstream approaches for few-shot text classification are based on meta-learning. 
In this section, we first introduce the background of meta-learning and then review how to apply meta-learning in few-shot text classification.

\subsection{Meta-Learning}
Meta-learning, 
also known as ``learning to learn'', 
refers to improving the learning ability of a model through multiple meta-tasks so that it can easily adapt to new tasks.
Existing approaches can be grouped into three main categories: 

\paragraph{Metric-based methods}
These kinds of methods~\cite{PROTO_snell2017prototypical,MeTNet_han2023meta}
aim to learn an appropriate distance metric to measure the distance between query samples and training samples. The label of a query sample is then predicted as that of the training sample with the smallest distance.
The representative methods include Siamese Network~\cite{SN_koch2015siamese}, Matching Network~\cite{vinyals2016matching_network}, PROTO~\cite{PROTO_snell2017prototypical} and Relation Network~\cite{Relation_network_sung2018learning}. 
Among these models, PROTO is simple-to-implement, fast-to-train and can achieve state-of-the-art results on several FSL tasks. 
Based on PROTO,
our proposed method is also a metric-based method.

\paragraph{Optimization-based methods}
These kinds of methods learn how to optimize.
Instead of simply using a traditional optimizer, such as stochastic gradient descent (SGD),
they train a meta-learner as an optimizer or adjust the optimization process.
A representative method is MAML~\cite{MAML_finn2017model}, 
which emulates the quick adaptation to unseen classes during the optimization process.
Other optimization-based models include Reptile~\cite{Reptile_nichol2018first},
iMAML~\cite{iMAML_rajeswaran2019meta} and MetaOptNet~\cite{MetaOptNet_lee2019meta}.


\paragraph{Model-based methods}

Model-based methods learn a hidden feature space
and predict the label of a query instance in an end-to-end manner, which lacks interpretability.
Compared with optimization-based methods, 
model-based methods could be easier to optimize but less generalizable to out-of-distribution tasks~\cite{metaSurvey:journals/corr/abs-2004-05439}.
The representative 
model-based methods include 
MANNs~\cite{MANN_santoro2016meta}, 
Meta networks~\cite{Meta_network_munkhdalai2017meta}, SNAIL~\cite{SNAIL_mishra2017simple} and CPN~\cite{CNP_garnelo2018conditional}.

\subsection{Few-Shot Text Classification}

Few-shot text classification
has received great attention recently~\cite{FSTC_example1:conf/eacl/SchickS21,MICK:conf/cikm/GengCZSZ20}.
In particular,
meta-learning has been applied to 
solve the problem~\cite{DBLP:conf/emnlp/GengLLZJS19,DS-FSL:conf/iclr/BaoWCB20}.
For example,
DS-FSL~\cite{DS-FSL:conf/iclr/BaoWCB20} adds distributional signatures 
(e.g. word frequency and information entropy) 
to the model within a meta-learning framework.
MEDA~\cite{DBLP:conf/ijcai/MEDA} jointly optimizes the ball generator and the meta-learner, such that the ball generator can learn to produce augmented samples that best fit the meta-learner.
MLADA~\cite{MLADA:conf/acl/HanFZQGZ21} integrates an adversarial domain adaptation network with a meta-learning framework to improve the model's adaptive ability for new tasks
and achieves the superior performance.
There also exist
methods that further extend
PROTO to the problem.
For example,
HATT-Proto~\cite{HATT:conf/aaai/GaoH0S19} 
learns weights of instances and features by
introducing the instance-level and feature-level attention mechanism, respectively.
LM-ProtoNet~\cite{LM:conf/cikm/FanBSL19} adds a triplet loss to PROTO
to improve the model generalization ability. 
IncreProtoNet~\cite{IncreProto:conf/coling/RenCCWL20} combines the deep neural network with PROTO to better utilize the training data.
LaSAML~\cite{DBLP:conf/acl/LaMSRA} improves performance of PROTO by incorporating label information into feature extractors.
LEA~\cite{hong2022lea} derives meta-level attention aspects using a new meta-learning framework.
ContrastNet~\cite{aaai2022_chen2022contrastnet} introduces a contrastive learning framework to learn discriminative representations for texts from different classes.
Despite the success,
all these methods disregard
the randomness of the sampled support sets when computing prototype vectors and 
employ a completely random construction of meta-tasks.
This adversely affects their wide applicability in various real-world tasks.

\section{Background}
In this section, we give a formal problem definition and summarize the standard meta-learning framework for few-shot classification~\cite{vinyals2016matching_network}.


\paragraph{Problem Definition}
Given a set of labeled samples from a set of classes $\mathcal{Y}_{train}$,
our goal is to develop a model that learns from these training data, so that we can make predictions over new (but related) classes, for which we only have a few annotations.
These new classes are denoted as $\mathcal{Y}_{test}$, which 
satisfies  $\mathcal{Y}_{train} \cap \mathcal{Y}_{test} = \emptyset$.

\paragraph{Meta-training} 
In meta-learning, we emulate the real testing scenario with meta-tasks during meta-training, so our model 
can learn to quickly adapt to new classes.
To create a training meta-task, 
we first sample $N$ classes from $\mathcal{Y}_{train}$.
After that,
for each of these $N$ classes, 
we sample $K$ instances as the support set $\mathcal{S}$ and $L$ instances as the query set $\mathcal{Q}$.
The support set is used as training data while the query set is considered as testing data.
Our model is updated based on the loss over these testing data.
Given the support set, we refer to the task of making predictions over the query set as \emph{$N$-way $K$-shot classification}.




\paragraph{Meta-testing}
In the testing stage, 
we also use meta-tasks to test whether our model can adapt quickly to new classes. 
To create a testing meta-task, 
we first sample $N$ new classes from $\mathcal{Y}_{test}$. 
Similar as in meta-training,
we then sample the support set and the query set from the $N$ classes, respectively. 
Finally,
we evaluate the average performance on the query set across all testing meta-tasks.

\section{Algorithm}
\label{sec_alg}

\begin{figure*}[t]
    \centering
    \includegraphics[scale=0.36]{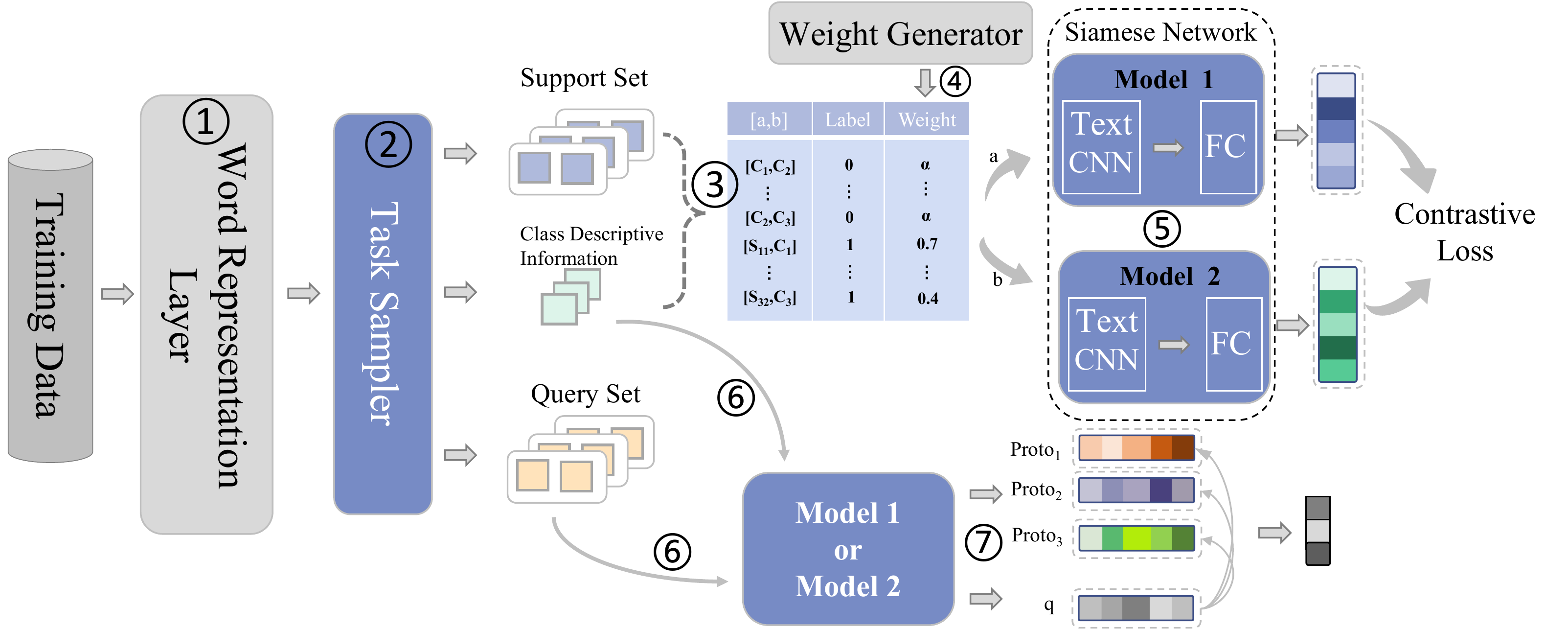}
    \caption{The overall architecture of \msn\ for a $3$-way $2$-shot problem. For details of each step, see Section~\ref{sec_alg}.}
    \label{fig:model}
\end{figure*}

In this section, we describe our \msn\ algorithm.
We first give an overview of \msn, which is illustrated in Figure~\ref{fig:model}.
\msn\ first represents each word with a $d$-dimensional embedding vector~(Step \ding{172}),
based on which the embeddings of samples and prototype vectors are initialized. 
Then it constructs meta-tasks by giving higher sampling probability to tasks that are hard to classify~(Step \ding{173}). 
After that, 
it generates sample pairs based on the support sets and prototype vectors~(Step \ding{174}), and also assigns weights to these sample pairs~(Step \ding{175}). 
\msn\ further employs 
a Siamese Network to map
samples and prototype vectors into a space that is much easier to be classified~(Step \ding{176}).
Finally, \msn\ feeds a query and all the prototype vectors into the Siamese Network~(Step \ding{177}) to derive their embeddings,
based on which the probability of the query in each class is calculated~(Step \ding{178}). 
The overall procedure of \msn\ is summarized in Algorithm~\ref{alg:Meta-SN}. Next, we describe each component in detail.


\subsection{Word Representation Layer}

In this layer, 
we 
use pre-training models,
such as fastText~\cite{DBLP:journals/corr/JoulinGBDJM16} and BERT~\cite{bert_devlin2018bert}, to
represent each word with a $d$-dimensional embedding vector. 
Then
for each class $c_j$,
we can construct its initial embedding vector
$f_0(c_j)$
by averaging the embeddings of words contained in its descriptive texts.
Similarly,
we can also derive the initial embedding vector $f_0(s_i)$ for a sample $s_i$.

\subsection{Task Sampler}

This module is used to construct meta-tasks from training data. 
Different from previous works that construct meta-tasks in a completely random manner, 
we assign higher sampling probability to tasks that are hard to classify.
Intuitively,
the closer the distance between prototype vectors, 
the more difficult the corresponding classes can be separated;
the more distant a sample in the support set is to the prototype vector,
the more difficult the task will be.
Therefore,
our goal is to construct more meta-tasks,
where prototype vectors are close to each other
and 
samples in a class
are far away from the corresponding  prototype vector.

To achieve the goal,
we first define a probability score $p_{i,j}^c$ to capture the correlation between classes $c_i$ and $c_j$:
\begin{eqnarray}   
\label{eq:pc}
p^c_{i,j} = \frac{e^{-dis({f_0}(c_i), {f_0}(c_j))}}{\sum^{|\mathcal{C}|}_{k=1}e^{-dis({f_0}(c_i), {f_0}(c_k))}},
\end{eqnarray}
where $|\mathcal{C}|$ is the number of classes in the training set 
and $f_0(\cdot)$ represents the initial embedding vector.


After that, 
given a class $c_i$,
we define a probability score $p_{ij}^s$ to sample the $j$-th instance $s_j$ in the class:
\begin{eqnarray}   
\label{eq:ps}
p^s_{i,j} = \frac{e^{dis({f_0}(c_i), {f_0}(s_j))}}{\sum^{m_i}_{k=1}e^{dis({f_0}(c_i), {f_0}(s_k))}},
\end{eqnarray}
where 
$m_i$ is the number of instances in class $c_i$.

Based on 
Equations~\ref{eq:pc} and~\ref{eq:ps},
we utilize a greedy algorithm to construct meta-tasks. 
We first randomly sample a class $c_1$ in $\mathcal{Y}_{train}$. 
Then we sample the second class $c_2$ with the probability scores
$\{p^c_{1,j}\}_{j=2}^{|\mathcal{C}|}$. 
From Equation~\ref{eq:pc}, 
a class that is closer to $c_1$ has a higher probability to be sampled. 
Next, we sample the third class $c_3$ based on the mean probability distribution $\{\frac{p^c_{1,j} + p^c_{2,j}}{2}\}_{j=3}^{|\mathcal{C}|}$, which indicates that a class with a small distance to both $c_1$ and $c_2$ will be more likely to be sampled.
We repeat the process to sample $N$ classes in total.
After $N$ classes are derived, 
for each class $c_i$, we 
constitute the support set $\mathcal{S}_i$ and the query set
$\mathcal{Q}_i$
by
sampling $K$ and $L$ instances 
via the probability distribution $\{p^s_{i,j}\}_{j=1}^{m_i}$, respectively.
From Equation~\ref{eq:ps},
a sample that is more distant from $c_i$ has a higher sampling probability. 
The pseudocode of meta-task sampling is summarized in Algorithm~\ref{alg:TS}.

\begin{figure}[!t]
  \begin{algorithm}[H]
    \caption{Task\_Sampler}
    \label{alg:TS}
    \begin{algorithmic}[1]
    \Require
      Training data $ \{\mathcal{X}_{train}, \mathcal{Y}_{train}\} $;
      $N$ classes in a meta-task;
      $K$ samples in each class in the support set and $L$ samples in each class in the query set.
      \Ensure a meta-task including a support set~$\mathcal{S}$, a query set~$\mathcal{Q}$ and a set~$\mathcal{C}_{target}$ of their label information.
      \State $\mathcal{S}, \mathcal{Q}, \mathcal{C}_{target} \leftarrow \emptyset, \emptyset, \emptyset$;
      \State $c_1 = \texttt{Sample}(Random, \mathcal{Y}_{train}, 1)$;
      \Comment $\texttt{Sample}(P, \mathcal{Y}, N)$ denotes selecting $N$ elements from $\mathcal{Y}$ with the probability $P$.
      \State $\mathcal{C}_{target} \leftarrow \mathcal{C}_{target} \cup c_1$;
      \For{$i \in [2, N]$}
        \State Calculate $P = \{ \frac{\sum_{k=1}^{i-1}p^c_{k,j}}{i-1} \}_{j=i}^{|\mathcal{C}|}$ by Equation~\ref{eq:pc}.
        \State $c_i = \texttt{Sample}(P, \mathcal{Y}_{train} \backslash \mathcal{C}_{target}, 1)$;
        \State $\mathcal{C}_{target} \leftarrow \mathcal{C}_{target} \cup c_i$;
      \EndFor
      \For{$c_i \in \mathcal{C}_{target}$}
        \State Calculate $P = \{p^s_{i,j}\}_{j=1}^{m_i}$ by Equation~\ref{eq:ps}.
        \State $\mathcal{S} \leftarrow \mathcal{S} \cup \texttt{Sample}(P, \mathcal{X}_{train}^{c_i}, K)$;  \Comment{$\mathcal{X}_{train}^{c_i}$ denotes samples with label $c_i$ in $\mathcal{X}_{train}$.}
        \State $\mathcal{Q} \leftarrow \mathcal{Q} \cup \texttt{Sample}(P, \mathcal{X}_{train}^{c_i} \backslash \mathcal{S}, L)$;
      \EndFor
      \State \Return
      $\mathcal{S}, \mathcal{Q}, \mathcal{C}_{target}$
    \end{algorithmic}
  \end{algorithm}
\end{figure}

\subsection{Constructing Sample Pairs}

After meta-tasks are sampled,
we next generate sample pairs.
For each meta-task, 
we first construct sample pairs from the
prototype vector set~${\Phi}$.
Specifically,
we pair prototype vectors $\phi_i$ with $\phi_j$, and denote the pair as $\left <\phi_i, \phi_j \right >$.
Further,
we pair each sample in the
support set~$\mathcal{S}$ with all the prototype vectors in~$\Phi$.
Specifically,
we pair sample $s_i$ with prototype vector $\phi_j$, and denote the pair as $\left <s_i, \phi_j \right >$.
If the two items in a pair are in the same class, 
we denote the label of the pair as 1; otherwise, 0. 
For each pair, we generate its weight by the \emph{weight generator} (see Section~\ref{sec:wg}).


\subsection{Weight Generator}
\label{sec:wg}

This module is used to learn weights for sample pairs.
For a sample pair 
$\left <s_i, \phi_j \right >$,
we define the weight of the sample pair to be inversely proportional to the average distance between $s_i$ and the query set $\mathcal{Q} = \{q_l\}_{l=1}^L$: 
\begin{eqnarray}   
\label{eq:sw}
w_{\left <s_i, \phi_j \right >} = \texttt{softmax}\left [-\frac{1}{L}\sum_{l=1}^{L}dis(f_{\theta}(s_i), f_{\theta}(q_l))\right ],
\end{eqnarray}
where 
$f_{\theta}(\cdot)$ is the embedding vector of a sample generated from the Siamese network (see Section~\ref{sec:SN}).
We use the \texttt{Softmax} function to normalize the weight over all the samples in the support set.
Further,
for a sample pair $\left <\phi_i, \phi_j \right >$,
we manually set its weight to a hyper-parameter $\alpha$,
which can be used to control
the distance between two different class prototype vectors.






\begin{figure}[!t]
  \begin{algorithm}[H]
    \caption{\msn\ Training procedure}
    \label{alg:Meta-SN}
    \begin{algorithmic}[1]
      \Require
      Training data $ \{\mathcal{X}_{train}, \mathcal{Y}_{train}\} $;
      $T$ meta-tasks and $ep$ epochs;
      $N$ classes in the support set or the query set;
      $K$ samples in each class in the support set and $L$ samples in each class in the query set;
      the model parameter $\theta$.
      \Ensure The model parameter $\theta$ after training.
      \State Randomly initialize the model parameters $\theta$;
      \For{each $i \in [1, ep]$}
      \For{each $j \in [1, T]$}
      \State $\mathcal{S}, \mathcal{Q}, \mathcal{C}_{target} \leftarrow \texttt{Task\_Sampler}(\mathcal{X}_{train}, \mathcal{Y}_{train}, N, K, L)$;
      \State Construct sample pairs by $\mathcal{S}, \mathcal{C}_{target}$;
      \State Calculate $w$ by Equation~\ref{eq:sw};
      \State Input sample pairs to the Siamese Network;
      \State Calculate $\mathcal{L}_{c}$ by Equation~\ref{contrastive_loss};
      \State Update $\theta$ to $\theta'$ by Equation~\ref{eq:my_maml_support};
      \State Input $\mathcal{Q}, \mathcal{C}_{target}$ to the model with parameter $\theta'$;
      \State Calculate $\mathcal{L}_{ce}$ by Equation~\ref{loss_ce};
      \EndFor
      \State Update $\theta$ by Equation~\ref{eq:our_maml};
      \EndFor
      \State \Return $\theta$
    \end{algorithmic}
  \end{algorithm}
\end{figure}

\subsection{Siamese Network}
\label{sec:SN}

Siamese network contains two identical sub-networks that
have the same network architecture with shared parameters to be learned.
Each sub-network consists of a TextCNN~\cite{TextCNN:conf/emnlp/Kim14} and a fully connected~(FC) layer.
In practice,
sample pairs are taken as the input of the Siamese network, where each sample is fed into a sub-network.


To optimize the Siamese network,
given a set of sample pairs $\{\left< x_{il},x_{ir} \right> \}_{i=1}^n$ with a label set $\{y_i\}_{i=1}^n$,
we utilize the contrastive loss function defined in Equation~\ref{contrastive_loss}, 
which aims to enlarge the distance between two samples in zero-labeled pairs and 
shortens that between two samples in one-labeled pairs.
\begin{eqnarray}
\label{contrastive_loss}
\mathcal{L}_{c}(\theta)\! &=&\! \sum_{i=1}^{n}w_{\left< x_{il},x_{ir} \right>} [y_{i}dis(f_{\theta}(x_{il}), f_{\theta}(x_{ir})) \nonumber\\
\! &+&\! (1-y_{i})max(0, \delta-dis(f_{\theta}(x_{il}), f_{\theta}(x_{ir}))) ],
\end{eqnarray}
Here, 
$\theta$ denotes the trainable parameters of the Siamese network. 
We also introduce a margin
$\delta$. 
For zero-labeled pairs,
they can only contribute to the loss function if their distance is smaller than the margin.
We 
update model parameters $\theta$ with SGD:
\begin{eqnarray}   
\label{eq:my_maml_support}
\theta'=\theta-\alpha\nabla_\theta \mathcal{L}_{c}(\theta),
\end{eqnarray}
where $\alpha$ is the learning rate.
With one-step update,
$\theta$ becomes $\theta'$.
Based on $\theta'$,
the Siamese network can map query instances and 
prototype vectors
into low-dimensional embedding vectors.
After that, 
we calculate the probability logits of a query $q_i$ in class $c_j$ 
and feed the results into a \texttt{cross-entropy} function:
\begin{eqnarray}   
\label{loss_ce}
\mathcal{L}_{ce}(\theta') = \sum_{i=1}^{L}-log(\frac{e^{-dis(f_{\theta'}(q_i), f_{\theta'}(c_j))}}{\sum_{k=1}^{N}e^{-dis(f_{\theta'}(q_i), f_{\theta'}(c_k))}}).
\end{eqnarray}
Following the optimization strategy in MAML~\cite{MAML_finn2017model},
we update $\theta$ by:
\begin{eqnarray}   
\label{eq:our_maml}
\theta \leftarrow \theta-\beta\nabla_\theta \frac{1}{T}\sum_{t=1}^{T}\mathcal{L}_{ce}(\theta'_t),
\end{eqnarray}
where $\beta$ is the meta learning rate and $T$ represents the number of meta-tasks in each epoch. 
This boosts the generalization ability of the model to unseen classes with only one-step update.
The overall procedure of \msn\ is summarized in Algorithm~\ref{alg:Meta-SN}.

\section{Experiments}

In this section, 
we comprehensively evaluate the performance of \msn.
In particular,
we compare the classification accuracy of \msn\ with seven other methods on six benchmark datasets to show the effectiveness of our model. 

\subsection{Datasets}

\begin{table}[t]
\centering
\caption{
Statistics of datasets.
}
\begin{tabular}{cccccc}
\hline
Dataset & $\texttt{\#}$ tokens$/$example & $\texttt{\#}$ samples & $\texttt{\#}$ train cls & $\texttt{\#}$ val cls & $\texttt{\#}$ test cls\\
\hline
HuffPost & 11 & 36900 & 20 & 5 & 16 \\
Amazon & 140 & 24000 & 10 & 5 & 9 \\
Reuters & 168 & 620 & 15 & 5 & 11 \\
20 News & 340 & 18820 & 8 & 5 & 7 \\
RCV1 & 372 & 1420 & 37 & 10 & 24 \\
FewRel & 24 & 56000 & 65 & 5 & 10 \\
\hline
\end{tabular}
\label{tab datasets}
\end{table}

We use six benchmark datasets:
{HuffPost}~\cite{dataset,Huffpost_misra2021sculpting}, 
{Amazon}~\cite{DBLP:conf/www/HeM16},
{Reuters-21578}~\cite{reuters:lewis1997reuters},
{20 Newsgroups}~\cite{DBLP:conf/icml/Lang95},
{RCV1}~\cite{RCV1:journals/jmlr/LewisYRL04}
and 
{FewRel}~\cite{FewRel:conf/emnlp/HanZYWYLS18}.
In particular,
the first five are for text classification while the last one is for few-shot relation classification.
Statistics of these datasets are summarized in Table~\ref{tab datasets}.
All processed datasets and their splits are publicly available.

\subsection{Experiment Setup}

\subsubsection{Baselines}
We compare \msn\ with seven state-of-the-art methods,
which can be grouped into three categories:
(1) \emph{metric-based methods}: PROTO~\cite{PROTO_snell2017prototypical}, HATT-Proto~\cite{HATT:conf/aaai/GaoH0S19}
and ContrastNet~\cite{aaai2022_chen2022contrastnet};
(2) \emph{optimization-based methods}: MAML~\cite{MAML_finn2017model};
and (3) \emph{model-based methods}: Induction Networks~\cite{DBLP:conf/emnlp/GengLLZJS19}, DS-FSL~\cite{DS-FSL:conf/iclr/BaoWCB20} and MLADA~\cite{MLADA:conf/acl/HanFZQGZ21}.
Specifically,
{HATT-Proto} extends PROTO by adding instance-level and feature-level attention to the prototypical network.
ContrastNet introduces a contrastive learning framework to learn discriminative representations for texts from different classes.
Induction Networks learns a class-wise representation by leveraging the dynamic routing algorithm in the meta-learning training procedure.
DS-FSL
is a model that
utilizes distributional signatures of words
to extract meta-knowledge.
{MLADA} 
integrates an adversarial domain adaptation network with a meta-learning framework
to improve the model's adaptability to new tasks.

\subsubsection{Implementation Details}

We implemented \msn\ by PyTorch. 
The model is initialized by He initialization~\cite{HE:conf/iccv/HeZRS15} and trained by Adam~\cite{Adam:journals/corr/KingmaB14}. We run the model with the learning rates 0.2 for contrastive loss and 0.00002 for cross-entropy loss on all the datasets.
We apply early stopping when the validation loss fails to improve for 20 epochs.
Since
ContrastNet 
adopts BERT as the pre-training model for word embeddings
while most other competitors 
like HATT-Proto, DS-FSL and MLADA
use fastText
in their original papers,
we implemented both fastText-based and BERT-based\footnote{We
use the pretrained \texttt{bert-base-uncased} model for all datasets.}
\msn\ for fair comparison.
In Siamese network, we follow  \cite{TextCNN:conf/emnlp/Kim14} 
to use the 1-dimensional filter of sizes [1, 3, 5], 
each with 16 feature maps in CNN. 
We set the dimensionality of the fully connected layer to 64 and
the number of meta-tasks $T$ in each epoch to 3. 
We also fine-tune $\alpha$ (weight of sample pairs composed of two prototype vectors)
by grid search over \{1, 3, 5, 7, 9\} 
and set it to 5 on all the datasets. 
For Induction Network and DS-FSL, 
we report their results from \cite{DS-FSL:conf/iclr/BaoWCB20}. 
For other competitors,
part of
their results
are derived from the original papers; for the datasets where results are absent,
we use the original codes released by their authors and fine-tune the parameters of the models.
We run all the experiments on a single NVIDIA v100 GPU.
In our experiments,
we set 
$K$ to 1 in 1-shot task, 5 in 5-shot task and 
$L$ to 25. 
We evaluate the model performance based on 1,000 meta-tasks in meta-testing and report the average accuracy over 5 runs.

\subsection{Classification Results}

\begin{table}[t]
	\caption{Mean accuracy (\%) of 5-way 1-shot classification and 5-way 5-shot classification over all the datasets. We highlight the best results in bold.}
	\begin{center}
	\resizebox{1\columnwidth}{!}{
		\begin{tabular}{c|ccccccccccccc|cc}
			\hline
			\hline
			&\multirow{2}{*}{Methods}&\multicolumn{2}{c}{HuffPost}&\multicolumn{2}{c}{Amazon}&\multicolumn{2}{c}{Reuters}&\multicolumn{2}{c}{20News}&\multicolumn{2}{c}{RCV1}&\multicolumn{2}{c}{FewRel}&\multicolumn{2}{c}{Average} \\
			\cline{3-16}
            &&1 shot&5 shot&1 shot&5 shot&1 shot&5 shot&1 shot&5 shot&1 shot&5 shot&1 shot&5 shot&1 shot&5 shot \\
            \hline
			\multirow{7}{*}{fastText-}&MAML~\cite{MAML_finn2017model}
			& 35.9 & 49.3 & 39.6 & 47.1 & 54.6 & 62.9 & 33.8 & 43.7 & 39.0 & 51.1 & 51.7 & 66.9 & 42.4 & 53.5  \\
			&PROTO~\cite{PROTO_snell2017prototypical}
			& 35.7 & 41.3 & 37.6 & 52.1 & 59.6 & 66.9 & 37.8 & 45.3 & 32.1 & 35.6 & 49.7 & 65.1 & 42.1 & 51.1  \\
			&Induct~\cite{DBLP:conf/emnlp/GengLLZJS19}
			& 38.7 & 49.1 & 34.9 & 41.3 & 59.4 & 67.9 & 28.7 & 33.3 & 33.4 & 38.3 & 50.4 & 56.1 & 40.9 & 47.6  \\
			&Hatt-Proto~\cite{HATT:conf/aaai/GaoH0S19}
			& 41.1 & 56.3 & 59.1 & 76.0 & 73.2 & 86.2 & 44.2 & 55.0 & 43.2 & 64.3 & 77.6 & 90.1 & 56.4 & 71.3  \\
			&DS-FSL~\cite{DS-FSL:conf/iclr/BaoWCB20}
			& 43.0 & 63.5 & 62.6 & 81.1 & 81.8 & 96.0 & 52.1 & 68.3 & 54.1 & 75.3 & 67.1 & 83.5 & 60.1 & 78.0  \\
			&MLADA~\cite{MLADA:conf/acl/HanFZQGZ21}
			& 45.0 & 64.9 & 68.4 & 86.0 & 82.3 & 96.7 & 59.6 & 77.8 & 55.3 & 80.7 & 81.1 & 90.8 & 65.3 & 82.8  \\
			&\textbf{\msn}& \bf54.7 & \bf68.5 & \bf70.2 & \bf87.7 & \bf84.0 & \bf97.1 & \bf60.7 & \bf78.9 & \bf60.0 & \bf86.1 & \bf84.8 & \bf93.1 & \bf69.1 & \bf85.2  \\
			\hline
			\multirow{2}{*}{BERT-}
			&ContrastNet~\cite{aaai2022_chen2022contrastnet}
			& 52.7 & 64.4 & 75.4 & 85.2 & 86.2 & 95.3 & 71.0 & 81.3 & 65.7 & 87.4 & 85.3 & 92.7 & 72.7 & 84.3  \\
			&\textbf{\msn}
			& \bf63.1 & \bf71.3 & \bf77.5 & \bf89.1 & \bf87.9 & \bf96.7 & \bf72.1 & \bf83.2 & \bf67.3 & \bf88.9 & \bf86.8 & \bf94.6 & \bf73.6 & \bf87.3  \\
			\hline
			\hline
		\end{tabular}
		}
		\label{tab main experiment}
	\end{center}
	
\end{table}

We report the results of
5-way 1-shot classification and 5-way 5-shot classification
in Table~\ref{tab main experiment}.
From the table,
\msn\ achieves the best results across all the datasets.
For example,
in the fastText-based comparison,
\msn\ achieves
an average accuracy of $69.1\%$ in 1-shot classification and $85.2\%$ in 5-shot classification, respectively.
In particular,
it outperforms the runner-up model MLADA by
a notable $3.8\%$ and $2.4\%$ improvement in both cases. 
When compared against the PROTO model, \msn\ leads by $27.0\%$ and $34.1\%$ on average in 1-shot and 5-shot classification, respectively. 
These results clearly demonstrate that our model is very effective in improving PROTO. 
While
Hatt-Proto upgrades PROTO by learning the importance of labeled samples,
it disregards the
randomness of the sampled support sets when computing prototype vectors and 
constructs meta-tasks 
randomly,
which degrades its performance.
Further,
in the BERT-based comparison,
\msn\ also outperforms ContrastNet over all the datasets.
All
these results show that
\msn,
which generates prototype vectors from external knowledge,
learns sample weights
and constructs hard-to-classify meta-tasks,
can perform reasonably well.

\subsection{Ablation Study}

\begin{table*}[t]
	\caption{Ablation study: mean accuracy ($\%$) of 5-way 1-shot classification and 5-way
5-shot classification over all the datasets. We highlight the best results in bold. All the results are based on fastText.}
	\begin{center}
	\resizebox{1\columnwidth}{!}{
		\begin{tabular}{ccccccccccccc|cc}
			\hline
			\hline
			\multirow{2}{*}{Models}&\multicolumn{2}{c}{HuffPost}&\multicolumn{2}{c}{Amazon}&\multicolumn{2}{c}{Reuters}&\multicolumn{2}{c}{20News}&\multicolumn{2}{c}{RCV1}&\multicolumn{2}{c}{FewRel}&\multicolumn{2}{c}{Average} \\
			\cline{2-15}
            &1 shot&5 shot&1 shot&5 shot&1 shot&5 shot&1 shot&5 shot&1 shot&5 shot&1 shot&5 shot&1 shot&5 shot \\
            \hline
			Meta-SN-rpv & 46.0 & 61.5 & 62.9 & 80.1 & 75.0 & 89.9 & 52.2 & 70.0 & 52.3 & 79.1 & 76.1 & 85.9 & 60.8 & 77.9  \\
			Meta-SN-ew & 51.4 & 64.9 & 68.4 & 83.3 & 81.1 & 93.4 & 57.9 & 74.4 & 57.7 & 82.5 & 81.7 & 89.3 & 66.4 & 81.3  \\
			Meta-SN-rts & 52.1 & 66.1 & 69.3 & 84.5 & 81.6 & 95.1 & 60.0 & 76.8 & 58.7 & 84.2 & 79.7 & 88.8 & 66.9 & 82.6  \\
            Meta-SN-ln & 53.8 & 68.0 & 69.6 & 87.1 & 83.3 & 96.0 & 59.8 & 78.0 & 59.5 & 85.4 & 84.3 & 92.6 & 68.4 & 84.5  \\
			\hline
			\textbf{\msn}& \bf54.7 & \bf68.5 & \bf70.2 & \bf87.7 & \bf84.0 & \bf97.1 & \bf60.7 & \bf78.9 & \bf60.0 & \bf86.1 & \bf84.8 & \bf93.1 & \bf69.1 & \bf85.2  \\
			\hline
			\hline
		\end{tabular}
		}
		\label{tab ablation}
	\end{center}
	
\end{table*}

We conduct an ablation study to understand the characteristics of the main components of \msn.
One variant 
ignores the randomness of the sampled support sets and
directly uses the mean embedding vectors of samples in the support sets as the prototype vectors.
We call this variant \textbf{\msn-rpv} (\textbf{r}andom \textbf{p}rototype \textbf{v}ectors).
To show the importance of weight learning for samples,
we set equal weights for all sample pairs and call this variant \textbf{\msn-ew} (\textbf{e}qual \textbf{w}eights).
Another variant 
removes the task sampler and constructs meta-tasks in a completely random way.
We call this variant \textbf{\msn-rts} (\textbf{r}andom \textbf{t}ask \textbf{s}ampler).
Moreover, 
to study how external knowledge affects the classification results, 
we only extract knowledge from label names to construct prototype vectors 
and call this variant \textbf{\msn-ln} (\textbf{l}abel \textbf{n}ames).
This helps to study the model robustness towards the richness of external descriptive information.
The results of the ablation study are reported in Table \ref{tab ablation}.
From the table,
we observe:
(1) 
\msn\ significantly outperforms 
\msn-rpv in all the comparisons across datasets.
This shows the importance of eliminating the adverse impact induced by the randomness of sampled support sets when calculating prototype vectors.
(2) 
\msn\ also beats \msn-ew clearly.
For example,
in 5 shot classification task on Amazon, the accuracy of \msn\ is $87.7\%$ while that of \msn-ew is only $83.3\%$.
This shows the importance of weight learning for sample pairs.
(3) \msn\ leads \msn-rts in all the 1-shot classification and 5-shot classification tasks.
This is because 
\msn-rts constructs meta-tasks randomly while \msn\ focuses more on the hard-to-classify meta-tasks to boost the model training.
(4)
The average performance gaps between 
\msn-ln and \msn\ on both 1-shot and 5-shot classification tasks are $0.7\%$.
This shows that the inclusion of external descriptive texts for class labels can bring only marginal improvement on the classification results.
Although we use only 
class names to derive the 
initial embeddings of class prototype vectors,
\msn\ can progressively refine these embeddings and thus lead to superior performance.


\subsection{Visualization}

\begin{figure*}[t]
    \centering
    \includegraphics[width=0.977\textwidth]{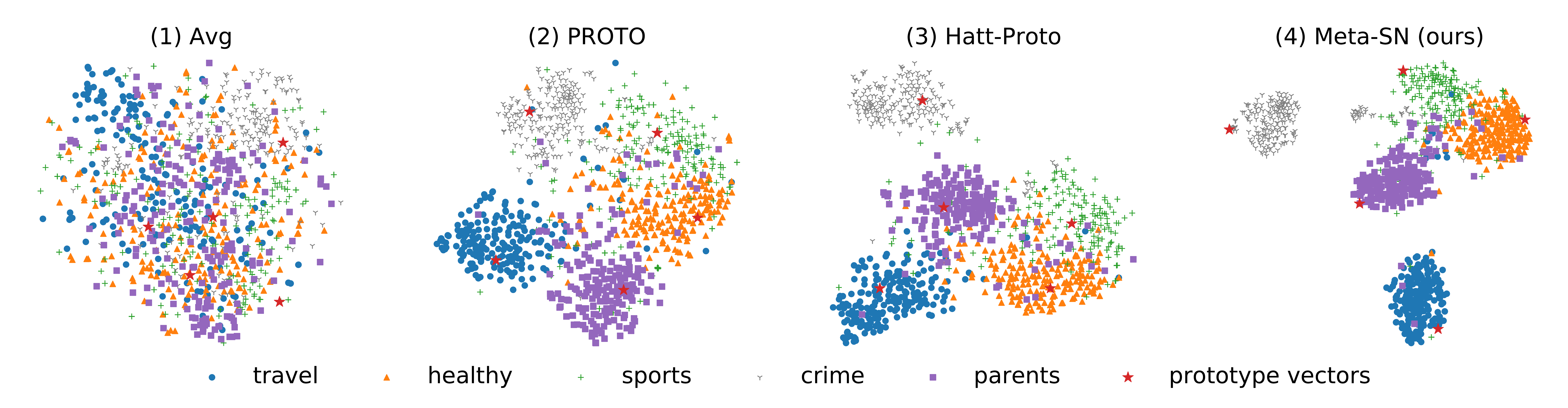}
    \caption{
    The t-SNE visualization comparison of sentence embeddings in meta-testing 
    on the HuffPost dataset.
    Note that these classes are unseen in meta-training.
    All the results are based on fastText.
    }
    \label{fig:analysis}
\end{figure*}

\begin{figure*}[t]
    \centering
    \includegraphics[width=0.93\textwidth]{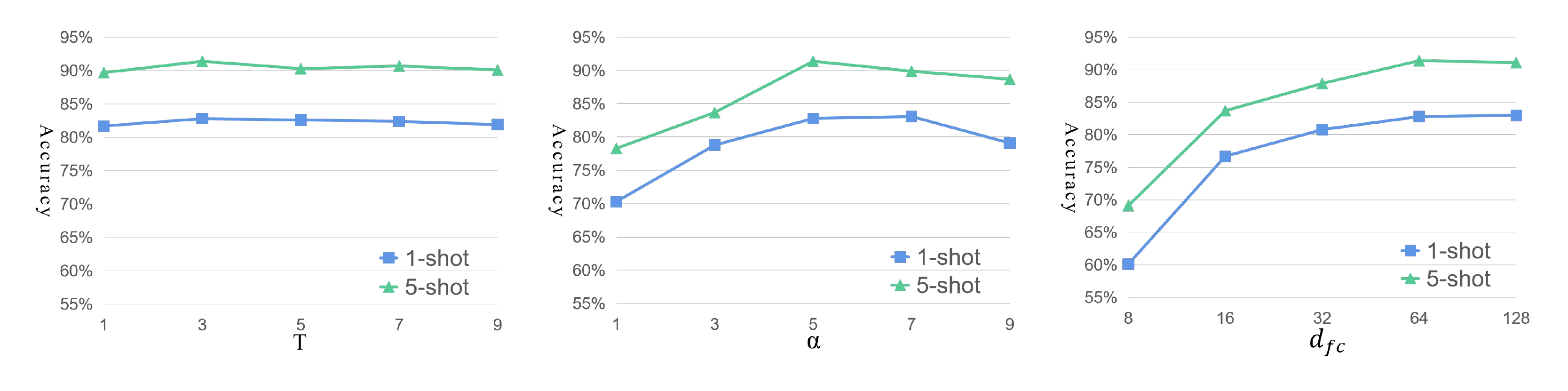}
    \caption{Hyperparameter sensitivity study on the HuffPost dataset.
    Here, $T$ is the number of meta-tasks in each training epoch, $\alpha$ is the weight controlling the distance between prototype vectors, and $d_{fc}$ is the output embedding dimensionality of query samples. All the results are based on fastText.}
    \label{fig:hyper}
\end{figure*}

We next evaluate the quality of 
generated embeddings of \msn.
Specifically,
Figure~\ref{fig:analysis} uses t-SNE~\cite{van2008visualizing} to visualize the sentence embeddings of the query set generated from different methods on the HuffPost dataset. 
For other datasets,
we observe similar results that are omitted due to the space limitation.

Figure~\ref{fig:analysis}(1)
shows the results of \emph{AVG},
which generates the sentence embedding by directly averaging the embeddings of the words contained in the sentence.
From the figure, 
embeddings of 
sentences with different class labels are entangled with each other.
While PROTO (Figure~\ref{fig:analysis}(2)) and Hat-Proto (Figure~\ref{fig:analysis}(3)) can produce higher quality of sentence embeddings,
it still fails to distinguish some classes.
Further,
our method \msn\ can generate embeddings that are clearly separated on all the datasets,
as shown in Figure~\ref{fig:analysis}(4).
These results show the superiority of \msn\ in generating high-quality embeddings.


\subsection{Hyper-parameter Sensitivity Analysis}
\label{ex:hyper}
We end this section with a sensitivity analysis on the hyper-parameters. 
In particular,
we study three main hyper-parameters:
the number of meta-tasks $T$ used in each training epoch,
the weight $\alpha$ controlling the distance between prototype vectors, and the output embedding dimensionality $d_{fc}$ of query samples.
In our experiments,
we vary one hyper-parameter with others fixed.
The results on the HuffPost dataset are shown in Figure~\ref{fig:hyper}. 
For other datasets,
we observe similar results, which are omitted due to the limited space.
From the figure, we see:

(1) \msn\ gives very stable performance over a wide range of $T$ values. This shows that \msn\ is insensitive to the number of meta-tasks in meta-training.

(2) With the increase of $\alpha$, 
the distance between prototype vectors of different classes is zoomed in and the performance of \msn\ generally becomes better. 
We also notice a mild dip when $\alpha$ is set too large. 
This is because a large value of $\alpha$ enforces the model to focus more on enlarging the distance between different class prototype vectors, but disregards the importance of shortening the distance between a sample and its corresponding class prototype vector.

(3) As the output embedding dimensionality $d_{fc}$ increases,
\msn\ achieves better performance.
This is because when the dimensionality is small, 
the embedding vectors cannot capture enough information for classification.

\section{Conclusion}

In this paper, 
we studied the few-shot text classification problem
and 
proposed a meta-learning Siamese network \msn. 
Based on PROTO,
\msn\
maps samples and prototype vectors of different classes into a low-dimensional space,
where
the inter-class distance between different prototype vectors is enlarged and the intra-class distance between samples and their corresponding prototype vectors is shortened. 
We generated prototype vectors based on the external descriptive texts of class labels instead of from the sampled support sets.
We learned the importance of samples in the support set based on their distances to the query set.
We also put forward a novel meta-task construction method, which 
samples more hard-to-classify meta-tasks to boost training.
We conducted extensive experiments to show that 
\msn\ can significantly outperform other
competitors on six benchmark datasets w.r.t. both text classification and relation classification tasks.
Future work includes applying \msn\ to other fields, such as computer vision, and exploring other representative meta-learning methods to improve their performance in few-shot text classification.

\section*{Acknowledgments}
This work has been supported 
by the National Natural Science Foundation of China 
under Grant No.
U1911203,
61977025
and
62202172.

\bibliographystyle{splncs04}
\bibliography{sample-base}

\end{document}